\documentclass[lettersize,journal]{IEEEtran}
\usepackage{fancyhdr,graphicx,amsmath,amssymb}
\usepackage[ruled,linesnumbered]{algorithm2e}
\usepackage{amsfonts}
\usepackage{subcaption}
\usepackage{array}
\usepackage{textcomp}
\usepackage{stfloats}
\usepackage{url}
\usepackage{verbatim}
\usepackage[noadjust]{cite}
\usepackage{import}
\usepackage{makecell}

\newcounter{alphsubsub}  

\newcommand{\Mycommand}[1]{}
\SetKwComment{Comment}{*/}{}

\begin{document}

\title{A Stereo Visual SLAM System Using Object-Level Motion Estimation and Geometric Filtering Based on Cross Disparity}

\author{Sujan Kumar Dhali, Bhaskar Dasgupta
\thanks{S. K. Dhali is with Department of Mechanical Engineering,
Indian Institute of Technology Kanpur, Kanpur 208016, India 
(e-mail: sujankd20@iitk.ac.in).}
\thanks{B. Dasgupta is with Department of Mechanical Engineering, Indian Institute of Technology Kanpur, Kanpur 208016, India 
(e-mail: dasgupta@iitk.ac.in).}}

\markboth{}%
{Shell \MakeLowercase{\textit{et al.}}: }


\maketitle

\begin{abstract}
This paper presents OCD SLAM, a dynamic stereo visual SLAM framework that extends ORB-SLAM2 by jointly addressing dynamic objects and dynamic features in the scene. Usual visual SLAM systems operating in dynamic environments often fail in the presence of moving objects, due to the static-world assumption used in pose estimation and mapping. To address this predicament, we introduce a novel geometric approach based on the discrepancy between disparity and a newly proposed notion called ``cross disparity", which exploits both temporal and stereo inconsistency to identify dynamic feature points. Complementary to this feature-level motion analysis, OCD SLAM integrates a 3D object detection module (SMOKE) with Kalman filter-based object tracking to perform object-level motion classification, enabling robust separation of static and dynamic scene elements for accurate pose estimation. The proposed approach has been evaluated on various sequences from the KITTI Odometry and KITTI Raw datasets. Results demonstrate that OCD SLAM achieves significant improvement in trajectory accuracy compared to ORB-SLAM2 and several state-of-the-art dynamic SLAM methods. Ablation studies further demonstrate the effectiveness of the cross disparity module in the KITTI Raw dataset and show that this method is able to detect dynamic features that are missed by the 3D object detection scheme alone.

\Mycommand{The proposed system integrates a 3D object detection module (SMOKE) to obtain object dimensions, orientation, and position, which are used for object tracking and velocity estimation using a Kalman filter under a constant-velocity motion model. Object-level motion labels are then assigned based on the estimated velocity magnitude and covariance. 
Additionally, this system introduces a geometric module based on the discrepancy between the disparity and the proposed cross disparity, which detects dynamic features belonging to objects not captured by the object detection module and improves the separation between static and dynamic feature points.} 
 
\end{abstract}


\section{Introduction}
\label{Introduction}
One of the fundamental requirements for applications like autonomous driving, robotic navigation etc., is the perception of the environment and self-localization. To address such purposes, simultaneous localization and mapping (i.e., SLAM) was developed. SLAM relies on onboard sensors such as LiDAR and cameras to estimate the vehicle’s trajectory while constructing a map of the environment. Visual SLAM (vSLAM) is a subset of the SLAM domain that uses cameras or camera-like sensors, and has become well known for its cost-effectiveness and strong performance. ORB-SLAM2~\cite{ORB-SLAM2}, DSO-SLAM~\cite{DSO}, LSD SLAM~\cite{LSD-SLAM}, RGBD SLAM~\cite{RGBD} are some of the well-known vSLAM algorithms that have been used extensively over the last decade. However, the key limitation of these methods is the assumption that the environment is static; thus, they fail when objects in the field of view are in motion.

To mitigate the effect of dynamic objects, several methods have been proposed, including geometric, deep learning, and hybrid approaches. In geometric methods, the motion or properties of image features are analyzed to segregate dynamic and static elements. These methods use several properties of the features, such as the epipolar or Euclidean distance between correspondences, the error between measured and projected depth, etc. In the case of deep learning methods, various learning models are used to detect predefined dynamic objects such as cars, bicycles, and pedestrians so that their influence can be excluded from the SLAM process. In other deep learning-based approaches, dynamic objects are identified by tracking their motion in the environment. These methods have several limitations, which are discussed in the following.

(a) In most deep learning-based methods, the segmented mask areas or the regions within the detected bounding boxes are removed from the frame so that features are not extracted from those regions. However, this may degrade performance, as it can remove areas that are temporarily static and useful for estimation, such as parked vehicles.

(b) Deep learning networks are typically trained to recognize a limited set of predefined object classes, such as cars or pedestrians. Consequently, if a dynamic object does not belong to any of these classes, the system may fail to detect it, and features on that object may negatively affect the bundle adjustment process. 

(c) On the contrary, geometric methods scrutinize each feature in the environment; often, through misinterpretation, removing many static features that clearly belong to static objects. 

(d) In geometric methods, epipolar or multi-view geometric constraints are commonly used, which mainly verify positional consistency (2D geometric relations) rather than depth consistency. Thus, these methods fail when dynamic objects move along the epipolar line, which frequently occurs in outdoor scenarios.

(e) Depth-based geometric methods also have limitations; they are highly sensitive to sensor noise, do not account for pixel-level accuracy, and often assume that 3D points are tracked perfectly.


\mbox{}

To solve the above problem, we propose a stereo SLAM system named OCD SLAM, which integrates a method for detecting and tracking 3D objects to identify moving and dynamic objects. Moreover, this paper introduces a novel geometric approach based on “cross disparity,” which effectively addresses disparity rather than depth, and provides a better detection method for motion classification of features. 

OCD SLAM is so named due to its being ``object-aware" through stereo vision and using cross disparity in the procedure. It is based on the framework of ORB-SLAM2. This method uses SMOKE (Single-Stage Monocular 3D Object Detection via Keypoint Estimation)~\cite{SMOKE} for detecting the 3D properties of objects, which are further used for tracking and later participate in the Kalman filter–based velocity estimation process that classifies the motion of the detected objects. After classification, the map points belonging to dynamic objects are excluded from the estimation process, whereas the features on static objects are labeled as static. To identify dynamic features that belong to the objects missed or undetected by the 3D segmentation module, the present paper introduces a novel geometric entity named cross disparity, enabling a better understanding of the motion properties of the feature. The geometric module based on it is further responsible for allocating higher weight to static points, which helps achieve better estimation in the bundle adjustment process. The proposed OCD SLAM is evaluated on public benchmark datasets, including the KITTI Odometry~\cite{KITTI} and KITTI raw~\cite{KITTI-RAW} sequences, and its performance is compared against ORB-SLAM2 and several state-of-the-art dynamic SLAM approaches.

The next section reviews the various methods of dynamic SLAM. Section~\ref{methodology} describes the methodology and implementation details. Section~\ref{Results} analyzes the experimental results and compares the proposed method with other algorithms. Finally, in Section~\ref{conclusion}, the method and results are summarized, outlining directions for future work.

\section{Related Work}
\label{Related work}
Literature on detection of dynamic points or features can be broadly categorized into two groups: geometric methods and deep learning-based methods.

(a) Geometric Methods: In geometric methods, the features of the environment are separated between static and dynamic points using various spatial and temporal properties. The geometric evaluation of features is particularly important in cases where no prior knowledge about the environment is available. Some of the geometric methods are discussed below.

Li and Lee~\cite{Li_Lee} proposed a method for computing visual odometry in dynamic environments by assigning higher weights to static features in the Intensity-Assisted Iterative Closest Points (IAICP) algorithm of Shile and Dongheui~\cite{IAICP}. The weights of the features (i.e., the detected foreground depth edges) were determined by comparing the ego-motion of the point clouds between the current and previous keyframes. MVS-SLAM~\cite{MVS_SLAM} used RANSAC based on the epipolar constraint to remove most outliers; furthermore, the differences in distance and angle between actual and reprojected 3D points were analyzed to detect the motion properties of these points. Similarly, Tan et al.~\cite{Tan} developed a method that used an improved RANSAC algorithm based on the inlier and outlier ratios of the previous frame for sample generation and hypothesis evaluation in the current frame. Dai et al.~\cite{Dai} proposed an RGB-D SLAM approach that removed dynamic points by optimizing the graph that represents the correlation between adjacent map points. Sun et al.~\cite{Sun} proposed a method for detecting dynamic regions using depth image segmentation and particle filter tracking. In StaticFusion~\cite{Staticfusion}, dynamic features were differentiated by comparing the geometric and photometric errors between the current and synthetically generated images. The synthetic image was produced using the estimated pose and the previous frame.

(b) Deep Learning Based Methods: 
Deep learning-based methods often offer an advantage over geometric methods as they leverage prior knowledge about the environment. This prior information about objects enables the accurate detection of entire dynamic regions, leading to more reliable segregation of dynamic points. Several studies have focused exclusively on deep learning-based approaches. Mask-SLAM~\cite{MaskSLAM} employs DeepLabv2~\cite{DeepLab} to segment predefined dynamic objects, using only the features from static regions for visual odometry estimation. Feng et al.~\cite{Fang} adopted a similar strategy but utilized an enhanced semantic segmentation network. Fangwei et al.~\cite{Fangwei}, Yubao et al.~\cite{RDS-SLAM}, and Ao et al.~\cite{DP-SLAM}  proposed methods that update the motion probability of features using purely semantic cues. Ganti et al.~\cite{Ganti} introduced a method that selects features from the environment with lower likelihoods of motion, improving robustness in dynamic scenes.

\mbox{}

The purely semantic or object detection-based methods have a major limitation, as they often fail to detect objects that are moving but not predefined as dynamic in the segmentation algorithm. Furthermore, whenever these objects are detected, they are interpreted as dynamic even if they are parked. To overcome this limitation, many approaches combine both deep learning and geometric techniques. This integration leverages the strengths of both domains—semantic awareness from deep networks and motion consistency from geometric constraints.

In DS-SLAM~\cite{DS_SLAM}, a semantic network such as SegNet~\cite{SegNet} is used to remove features from the predefined dynamic segments. The remaining features are then verified using the epipolar constraint for motion consistency. Similarly, the method proposed by Wen et al.~\cite{Shuhuan} uses SegNet, where static and dynamic objects are first segmented. The 3D scene flow of all features between consecutive frames is then computed, and dynamic features are identified by finding the discrepancy in the pixel displacement. DynaSLAM~\cite{Dyna_SLAM} uses Mask R-CNN~\cite{MaskRcnn} as its semantic network and computes the difference between projected and actual 3D points to determine the motion properties of map points. Dhali and Dasgupta ~\cite{Trifocal} proposed a better motion constraint based on the trifocal constraint and further incorporated YOLOv8~\cite{YOLOv8} as a segmentation module. Similar to Dai et al.~\cite{Dai}, AirDos~\cite{AirDOS} incorporates a rigidity constraint in addition to reprojection and motion constraints. This rigidity constraint ensures that the distance between two features of the same object remains consistent over time.

DGS-SLAM~\cite{DGS-SLAM} uses semantic segmentation with YOLACT++~\cite{YOLACT++} to remove predefined dynamic objects. It further segments image regions by applying k-means clustering on the depth map and determines the motion characteristics of each segment by counting the number of dynamic points identified through differences between actual and projected depths. RSO-SLAM~\cite{RSO_SLAM} is another method that uses k-means clustering on dense optical flow to segment regions with different motion patterns. Furthermore, it uses YOLOv5~\cite{YOLOv5} for semantic segmentation. Yang et al.~\cite{YANG} proposed a method that detects static points on dynamically labeled objects. They used DeepLabv3~\cite{DeepLabv3} to assign semantic labels to each feature and S2R-DepthNet~\cite{DepthNet} to compute feature depth. By jointly verifying the label and depth consistency, static points were identified. 

Several deep learning-based approaches detect dynamic objects by tracking them over time. Quang et al.~\cite{DRV-SLAM} introduced a method in which YOLOv8 is used to obtain both bounding boxes and segmentation masks of objects, then the bounding boxes of the objects are tracked through feature-based association to identify whether they are moving or static. Zhang et al.~\cite{VDO-SLAM} proposed a similar tracking strategy but further improved robustness by estimating object velocities from the associated 3D points. Song et al.~\cite{DGIM-VINS} proposed a dynamic SLAM framework that identifies and tracks objects using CenterNet~\cite{CenterNet}, which tracks objects based on the predicted displacement of object centers.

In OCD-SLAM, the focus is on object-level motion detection, where detected objects are tracked across consecutive frames, and stationary objects are utilized in the pose estimation process. To further enhance the system’s robustness, a novel geometric technique, termed cross disparity, is introduced in the present paper. This method enables feature-level motion classification by exploiting disparity consistency across stereo pairs, thereby facilitating more accurate separation of static and dynamic features.

\section{Methodology}
\label{methodology}
\begin{figure*}[hbtp]
\centering
  \includegraphics[width=1.45\columnwidth]{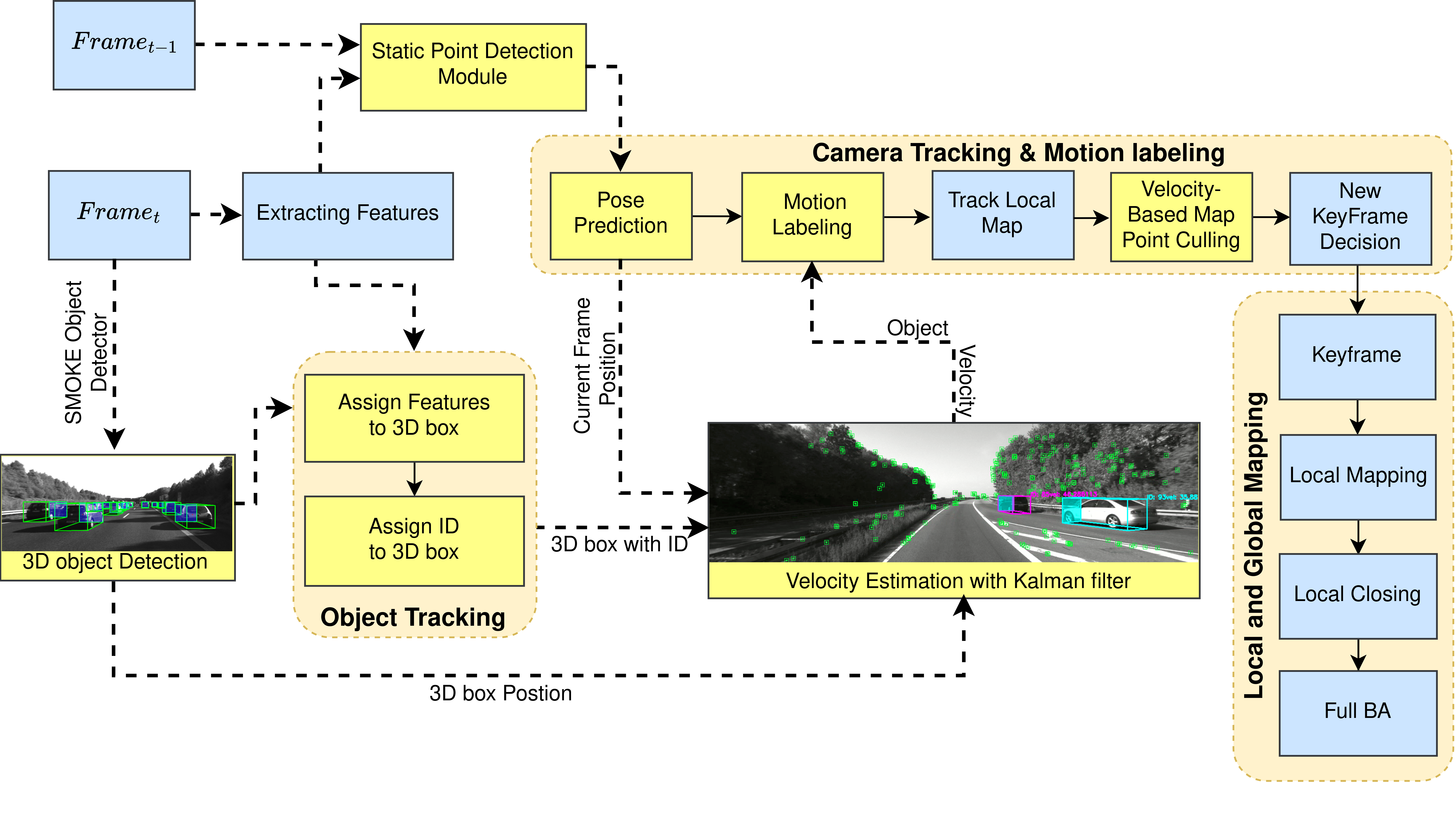}
  \caption{Framework of the OCD SLAM system. ORB SLAM2 framework is enhanced with several modules~(shown in yellow); namely static point detection, 3D object detection and tracking, motion labeling, and velocity-based map point culling.}
  \label{fig:Framework}
\end{figure*}
The proposed OCD-SLAM is based on the stereo architecture of ORB-SLAM2~\cite{ORB-SLAM2}, with the addition of a 3D object segmentation thread and various submodules. Figure~\ref{fig:Framework} illustrates the overall framework of OCD SLAM.

Our method consists of four parallel threads, namely tracking, local mapping, loop closing, and a 3D object detection thread. One image (by practice, usually the left) from each stereo pair (grayscale or RGB) first enters both the 3D object detection thread and the feature detection module in the tracking thread, where the former is responsible for estimating the 3D information of visible objects, such as their spatial dimensions, orientations, and 3D locations with respect to the camera coordinate system, while the latter detects image features. Next, the 3D information of the detected objects at several instances and the extracted features are passed through various submodules to estimate the velocity of the objects.

To estimate object velocity, the features detected by the ORB extractor are first associated with each detected object. These objects are then assigned unique IDs and tracked across consecutive frames, and their velocities are estimated using a Kalman filter. The estimated velocity is then used to assign a motion label to each object. Based on the velocity of the tracked object and its corresponding covariance, an object is classified as “Static” (or parked) or “Dynamic” (moving). After motion labels are assigned, they are used for filtering the features — features belonging to “Dynamic” objects are labeled as dynamic, while features associated with “Static” objects are treated as static.

Additionally, to identify dynamic features belonging to undetected moving objects, a geometric approach based on \textbf{cross disparity} is proposed. The static features determined through both velocity-based and geometry-based approaches are subsequently used in bundle adjustment. In the following sections, these components are explained in detail.

\subsection{Cross disparity-based geometric motion constraint}
To classify the motion properties of feature points (or associated map points), several geometric approaches have been proposed as discussed in Section~\ref{Related work}. In this work, a novel geometric approach is introduced that distinguishes static from dynamic features by comparing their disparity and cross disparity. 

In stereo vision, disparity refers to the horizontal shift of a feature’s projection between the left and right images of a rectified stereo pair. It is defined as
\begin{equation}
   D^{t} = (p^{t}_L)_x-(p^{t}_R)_x 
\end{equation}
where $p^{t}_L$ and $p^{t}_R$ denote the corresponding feature positions in the left and right images at time $t$, and $(\cdot)_x$ denotes the horizontal image coordinate. This disparity is commonly used to compute depth and reconstruct 3D map points.

We define \textbf{cross disparity} as follows. Let $p^{t-1}_L$ be a feature point in the previous left image, which reappears in the current left image at location $q^{t}_L$, and its corresponding 3D position is $Q_t$. When $Q_t$ is projected into the right image of the previous frame, its position is denoted by $q^{t-1}_R$. 
For a static point, $q^{t-1}_R$ is coincident with $p^{t-1}_R$, though not necessarily for a moving point.  
In the definition of disparity, using $q^{t-1}_R$ instead of $p^{t-1}_R$ would provide a new entity, named here as \textbf{cross disparity}, and given by 
\begin{equation}
D^{t-1}_{cross} = (p^{t-1}_L)_x-(q^{t-1}_R)_x \, .    
\end{equation}
The difference between the disparity $D^{t-1}$ and the cross disparity $D^{t-1}_{cross}$ of a feature indicates that the feature lies on a moving object, and it is therefore classified as dynamic. The converse of it, when these quantities are consistent (i.e., their values are close within a tolerance), the feature being static, is also mostly true.
The cross disparity constraint associated with the geometric error of a feature is defined by 
\begin{equation}
|D^{t-1}_{cross} - D^{t-1}|\leq \tau \; ,  
\end{equation}
using a tolerance $\tau$.

\begin{figure}[htp]
    \centering
    \includegraphics[width=0.8\linewidth]{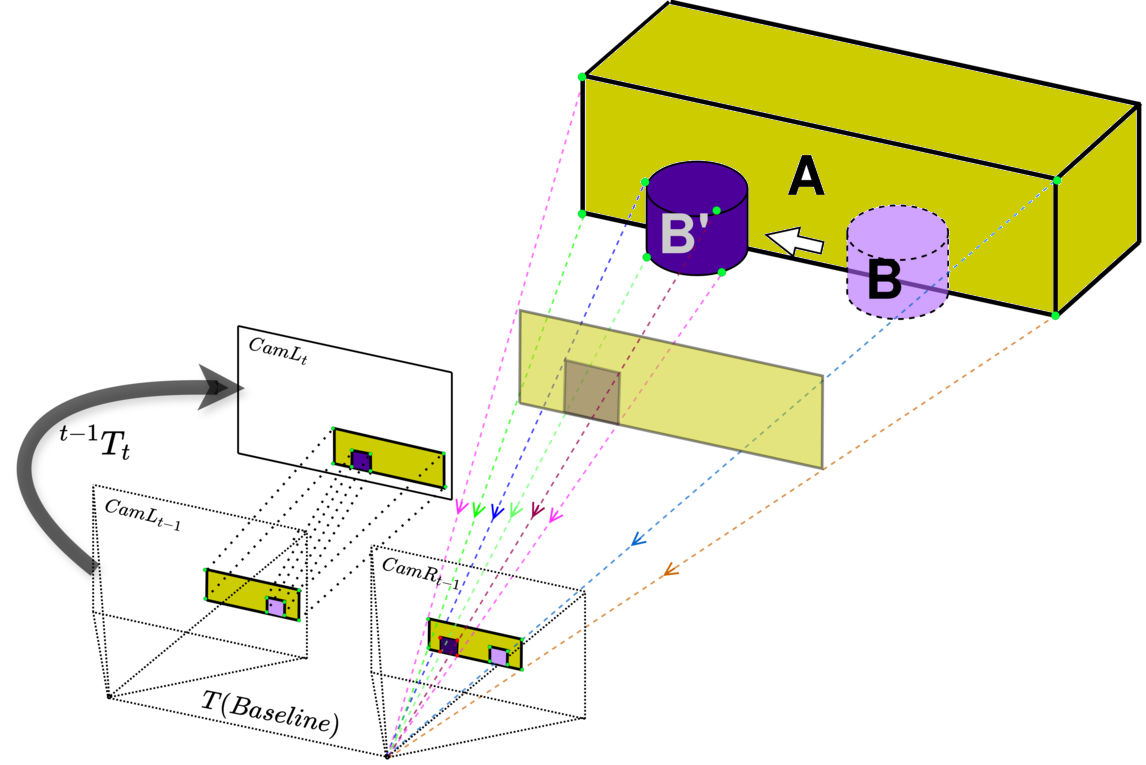}
    \caption{Principle of the proposed cross disparity constraint. Consistent disparity and cross disparity values indicate static features (object \textit{$A$}), while large deviations signify dynamic features (object \textit{$B$} → \textit{$B'$}) resulting from independent motion between frames.}
    \label{fig:cross_D}
\end{figure}

Figure~\ref{fig:cross_D} illustrates the principle of the proposed geometric method for distinguishing static and dynamic features. At time $(t-1)$, the stereo pair ($CamL_{t-1}$, $CamR_{t-1}$) captures both objects $A$ and $B$. Between time $(t-1)$ and $t$, the camera undergoes a known motion $^{t-1}T_t$, while object $A$ remains fixed and object $B$ moves to a new position $B'$. Consequently, in the current frame $CamL_{t}$, the image reflects the new scene configuration: $A$ in its original location and $B$ shifted to $B'$. 

To evaluate feature consistency, correspondences are first established between $CamL_{t-1}$, $CamL_{t}$ (shown in the blue dotted line), showing where the features observed in the previous frame appear in the current frame. These features of $CamL_{t}$ are then triangulated to obtain the 3D position, and projected back into the view of $CamR_{t-1}$ using the relative pose $^{t-1}T_t$ and baseline distance. For the static object $A$, the back-projected points align consistently with their locations in $CamR_{t-1}$, resulting in the disparity (between $CamL_{t-1}$, $CamR_{t-1}$) and the cross disparity (between $CamL_{t-1}$ and the back-projected points in $CamR_{t-1}$) being almost same. 

In contrast, features belonging to the moving object $B$ do not align. Although the features of object $B$ are observed in $CamL_{t-1}$, their corresponding 3D map points derived from $CamL_t$ project to different positions in $CamR_{t-1}$, leading to a mismatch between disparity and cross disparity. This difference in values shows that the feature lies on a dynamic object. Unlike conventional depth comparison methods, which validate depth consistency by projecting features from the current frame ($CamL_{t}$) onto the previous left image ($CamL_{t-1}$), the proposed cross disparity approach performs validation through an additional viewpoint ($CamR_{t-1}$), that means, checking geometric consistency from a completely different viewpoint. This cross-view verification establishes a wider geometric baseline that combines both temporal and stereo perspectives. 

Algorithm~\ref{ALGO} summarizes this method for identifying static points.
\begin{algorithm}[h]
\scriptsize
\caption{Cross Disparity Based Dynamic Feature Detection}
\label{ALGO}
\SetAlgoLined
\SetAlgoNoLine
\KwIn{Stereo pairs$(CamL_{t-1}, CamR_{t-1})$, $(CamL_t, CamR_t)$;
      Relative pose $T_{t-1 \rightarrow t}$; Disparity maps $D_{t-1}$}
\KwOut{Labels of static and dynamic feature points}
Detect and match feature correspondences between $CamL_{t-1}$ and $CamL_t$~\;
\For {each matched feature $p^{t-1}_L \leftrightarrow q^{t}_L$}
{
    a. Triangulate 3D point $Q^t$ from the current stereo pair $(CamL_t, CamR_t)$ \; 
    b. Back-project the transformed 3D point $Q^t$ into $CamR_{t-1}$ to obtain its projected position $q^{t-1}_R$\;
    c. $D^{t-1} = (p^{t-1}_L)_x - (p^{t-1}_R)_x$ \Comment*{Compute disparity}
    d. $D^{t-1}_X = (p^{t-1}_L)_x - (q^{t-1}_R)_x$ \Comment*[r]{Compute cross disparity}
    e. $\Delta D = |D^{t-1} - D^{t-1}_X|$ \Comment*[r]{Calculate C-disparity error}
    \eIf{$\Delta D \leq \tau$}{
        Label feature as \textit{static}\;
    }{
        Label feature as \textit{dynamic}\;
    }
}
\end{algorithm}

After the map points are categorized, separate weights are assigned to static and dynamic features, which are utilized in the bundle adjustment process. Notably, the geometric module operates only on features that are not part of detected objects (as described in Section~\ref{3D Object Detection Thread}), thereby avoiding the unintended removal of static features. In the proposed approach, static and dynamic points are assigned weights of 1.0 and 0.001, respectively. By assigning a significantly lower weight to dynamic points, their influence on pose estimation and map optimization is effectively minimized, thereby achieving the intended objective of the method.

\subsection{3D Object Detection Thread}
\label{3D Object Detection Thread}
Our system uses SMOKE (Single-Stage Monocular 3D Object Detection via Keypoint Estimation)~\cite{SMOKE} to estimate the 3D bounding box of objects. This object detection thread operates on each incoming image and produces the components of a 3D bounding box, such as a projected 3D keypoint $(x, y, z)$, dimensions $(h, w, l)$, and the yaw angle $(\varphi)$. 

SMOKE is trained on the KITTI~\cite{KITTI} dataset, which includes cars, pedestrians, and bicycles — the majority of dynamic objects in outdoor driving environments. As a backbone, this model uses DLA-34~\cite{DLA-34} with deformable convolutions to extract feature maps. It has two parallel prediction heads attached to such a feature map: (a) a keypoint classification head that predicts a heatmap whose peaks represent keypoint, i.e. the projected 3D centers of objects on the image plane, and (b) a regression head that encodes an 8-dimensional vector (depth offset, two subpixel offsets on the downsampled grid, residuals of object dimensions, and a vectorial encoding of the local orientation; i.e. $\bigl[\delta z,[\delta x,\delta y],[\delta h,\delta w,\delta l],[\sin{\alpha},\cos{\alpha}]\bigr]$) for each keypoint, which are required to construct 3D bounding box. In this method, the yaw angle is not directly estimated but can be computed using object location and local orientation as 
\begin{equation}
\label{eq:1}
\varphi =  \alpha + \arctan\left({\frac{x}{z}}\right) \; .
\end{equation}

\begin{figure}[htp]
\centering
  \includegraphics[width=\columnwidth]{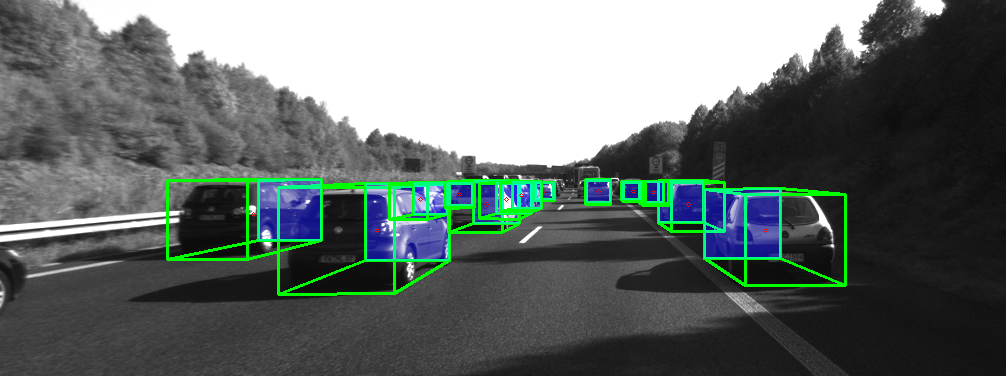}
  \caption{Visualization of 3D object detection results generated by SMOKE~\cite{SMOKE}. Each green box represents a projected 3D bounding box estimated from monocular images.}
  \label{fig:3D_box}
\end{figure}

With the dimension of an object, yaw angle ($\varphi$), and the 3D location of the center of the object, the eight corners of the 3D bounding box with respect to the camera coordinates can be computed using equation~\eqref{eq:2}. The eight corner points of the 3D bounding box can be projected further onto the image plane using equation~\eqref{eq:3} with the known camera calibration matrix. Here, $[\frac{u_i}{w_i}, \frac{v_i}{w_i}]^T$ is the image point of corner $\{ i \}$ and $K$ is the $3\times3$ calibration matrix. Figure~\ref{fig:3D_box} shows the projected 3D bounding box. 
\begin{align}
\begin{bmatrix} x^i_c\\ y^i_c\\ z^i_c \end{bmatrix}_{\{i=1:8\}} &=  \mathrm{Rot}(y,\varphi) \begin{bmatrix}\pm h/2\\ \pm w/2\\ \pm l/2 \end{bmatrix} + \begin{bmatrix} x\\ y\\ z \end{bmatrix} \label{eq:2} \, ,\\ 
\nonumber \\ 
\begin{bmatrix} u^i\\ v^i\\ w^i\end{bmatrix} &= K \begin{bmatrix} x^i_c\\ y^i_c\\ z^i_c \end{bmatrix} \label{eq:3} \, .
\end{align}
\subsection{Assigning features to each object}
\begin{figure}[htb]
\centering
    \subfloat[]{\label{fig:2d}{\includegraphics[width=0.205\textwidth]{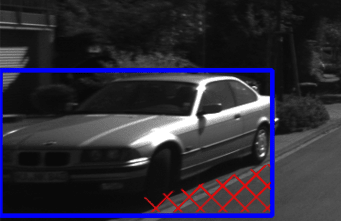}}} \hspace{5mm}
    \subfloat[]{\label{fig:hull}{\includegraphics[width=0.198\textwidth]{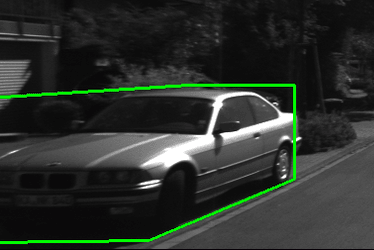}}}
  \caption{Illustration of object boundary refinement in OCD SLAM. The 2D bounding box (left) encloses a larger background area, while the constructed convex hull (right) defines a more accurate polygonal contour around the object.}
\end{figure}

For assigning a feature to an object, the first objective is to find a polygon that encloses that object, with minimum background inclusion. The usual practice of using the rectangular bounding box, as shown in Figure~\ref{fig:2d}, often encloses a significant area of the background in it.  In our method, we compute a convex hull with the projected corner points, which creates a polygon around the object as shown in Figure~\ref{fig:hull}. 
If any features from the image are inside the polygon, they are considered as a part of the object. One of the advantages of using a convex hull instead of a 2D bounding box is that the polygon computed from the projected 
3D corner points define a somewhat better (tighter) periphery of an object compared to the 2D bounding box. This more accurate coverage helps in producing clearer separation between background and foreground, and thereby helps in better allocation of feature points to an object.     
\subsection{Assigning ID to each object}
To track an object through the stream of frames, first we have to identify the object in both the current and the previous frames. After the data association is complete, each object is assigned an ID, based on its previous frame ID,  which helps it to be tracked in latter frames. Objects from the previous and current frames are associated by comparing them against multiple criteria. One method is to match the descriptors of the features associated with two objects. Qiang~et~al.~\cite{DRV-SLAM} used a similar approach for identifying these associations. However, in outdoor scenarios where cars are parked in proximity, comparing only the features is not good enough for association. 

Therefore, we have used additional properties for matching the object between two frames. Two such properties are (i) the relative difference between the volumes of two objects, where the volume can be computed by multiplying the estimated height, width, and length; and (ii) the difference between the estimated 3D locations of two objects. The first property checks the size difference between two objects, while the latter checks whether the objects are in sufficient positional continuity across the images. After computing these two similarity measures in addition to the feature matching, a weighted sum of these parameters is calculated for each object in the current frame with every other object of the previous frame, and the object corresponding to the lowest error is assigned the same ID as the previous object ID. Table~\ref{tab:tracking_eval} shows a quantitative comparison of object tracking performance between the feature-only method and the proposed method. The table reports, for each object, the total number of frames during which the object was detected by the object detection module, the number of frames in which it was correctly tracked with a consistent ID, and the number of inconsistent ID observed. Our proposed method has fewer to no such cases of ID inconsistency compared to the feature-only method, which indicates better tracking. 
\begin{table}[ht]
  \centering
  

  \subfloat{
    \renewcommand{\arraystretch}{2.5}
	\begin{tabular}{| p{7em} | c  c  c  c |}
		\hline
 		Object & Car A & Car B & Car C & Car D\\ \cline{1-5}
 		Frames Visible & 9 & 26 & 38 & 25\\ \cline{1-5}
 		Frames tracked correctly (Feature-only) & 8 & 22 & 25 & 16\\ 			\cline{1-5} 
 		Frames tracked correctly (Proposed) & \textbf{9} & \textbf{26} & 		\textbf{38} & \textbf{25} \\ \cline{1-5}
 		Different-ID  (Feature-only) & 1 & 4 & 15 & 9\\ \cline{1-5}
 		Different-ID  (Proposed) & \textbf{0} & \textbf{0} & \textbf{2} & 		\textbf{0}\\  \hline
	\end{tabular}
    \label{tab:tracking_comparison}
  }
  \hfill
  \subfloat{
    \includegraphics[width=0.45\textwidth]{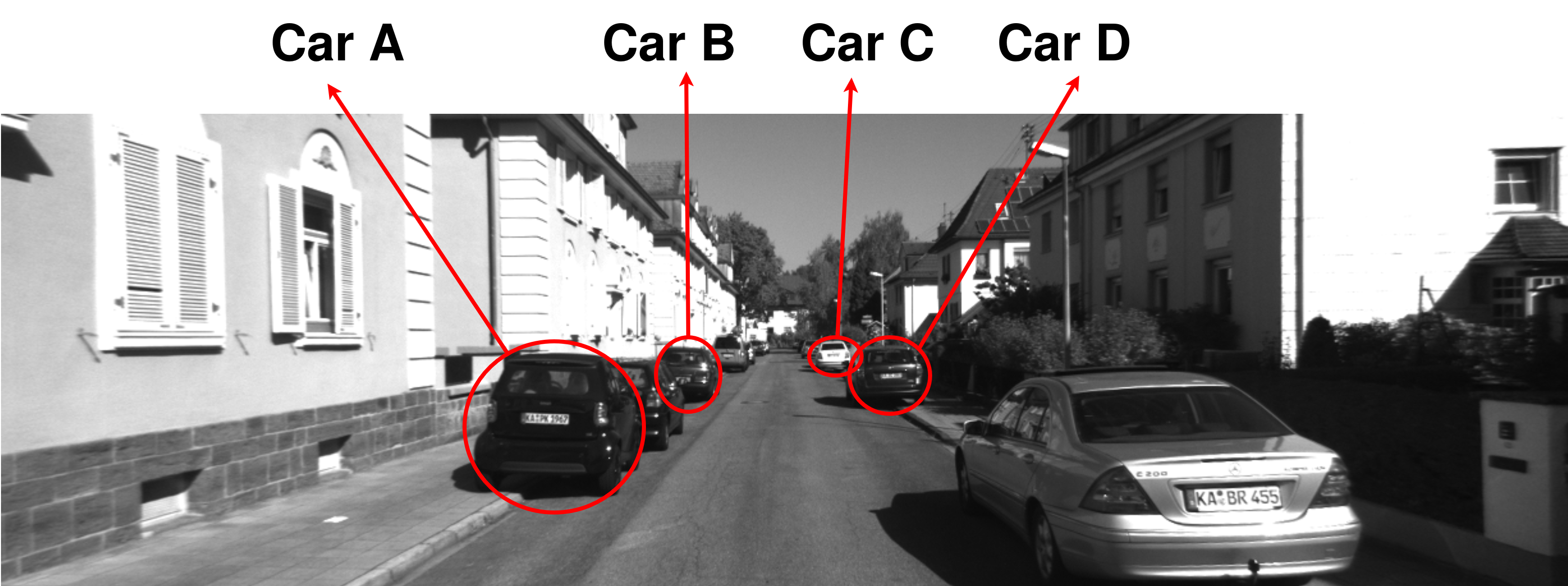}
    \label{fig:labeled_objects}
  }
  
  \caption{Quantitative comparison of object tracking performance between the feature-only and proposed method. The proposed method achieves significant reduction in cases of wrong ID assignments and improved tracking consistency across objects A–D.}
  \label{tab:tracking_eval}
\end{table}

If an object detected in the current frame has no predecessor or failed to find the correct data association in the previous frame, it is assigned a new unique ID.

\subsection{Velocity estimation of the tracked objects}
\label{Velocity estimation of tracked object}
This section describes how the velocity of each object is estimated and how motion labels are assigned accordingly. The procedure comprises two stages. 

(a) Velocity estimation with a Kalman filter: Although a naive idea of velocity can be obtained from finite difference of two consecutive positions, such estimates are highly sensitive to noise in both the 3D object location and the camera poses. To obtain reliable velocities, a Linear Kalman filter is employed, which takes into account the position of the object across all frames and updates the velocity iteratively.
In the Kalman filter, the state vector is defined as
\begin{equation}
X_{k}=\begin{bmatrix}
    p_k \\
    v_k
\end{bmatrix}.
\end{equation}
where $p_k$ is the 3D position of the object, transformed into the world frame, and $v_k$ is the object velocity. Under the assumption of the velocity $v_{k-1}$ and the sampling interval $\Delta t$ held constant, the state vector will become    
\begin{align}
&X_{k} = F X_{k-1}+w_k \, , \\
&F =
\begin{bmatrix}
1 & 0 & 0 & \Delta t & 0 & 0\\
0 & 1 & 0 & 0 & \Delta t & 0\\
0 & 0 & 1 & 0 & 0 & \Delta t\\
0 & 0 & 0 & 1 & 0 & 0\\
0 & 0 & 0 & 0 & 1 & 0\\
0 & 0 & 0 & 0 & 0 & 1 
\end{bmatrix}\, ;\nonumber
\end{align}
where $F$ is the state transition model, $w_k\sim\mathcal{N} (0,\,Q_k)$ is a Gaussian white noise with zero mean covariance $Q_k$. 

Since only the position is used as measurements, the \emph{observation model} is:
\begin{align}
&z_k = H X_{k}+n_k, \\
&H =
\begin{bmatrix}
1 & 0 & 0 & 0 & 0 & 0\\
0 & 1 & 0 & 0 & 0 & 0\\
0 & 0 & 1 & 0 & 0 & 0
\end{bmatrix}\, ;\nonumber
\end{align}
where $z_k$ is the observed position at time $k$, $H$ is the observation model, and $n_k\sim\mathcal{N} (0,\,R_k)$ is the measurement noise. 
The upper bound of the process noise and the measurement noise are set to $Q_k = 0.1*I_{6x6}$ and $R_k = 0.01*I_{3x3}$, respectively. The values of the bounds are empirically tuned to balance smoothness and responsiveness. Initial covariance of the velocity\Mycommand{, $P_{4:6,4:6}$,} is set high as the velocity is unknown and the initial velocity is set to zero. After several updates, once it becomes sufficiently small (indicating that the measured velocity has higher confidence), the velocity component $X_{4:6}$ is used for the assessment.

(b) Motion labeling: 
Each object is assigned any one of three different labels based on the estimated velocity and its uncertainty:
\begin{itemize}
\item \textbf{Static (Parked)} if the speed is below a fixed threshold;
\item \textbf{Dynamic (Moving)} if the speed exceeds the threshold;
\item \textbf{Unknown} if the velocity covariance is high (low confidence), or the object has just been assigned a new ID.
\end{itemize}
Features (i.e., the associated map points) are removed from the object when it is assigned a label, Dynamic.  For objects labeled \emph{Unknown}, though the map points are not removed, they are not used in the BA process. Finally, if the tracked object has a status of ``Parked", all the map points within it will be considered static and given higher weight. This step also updates the weight of the map point, which was accidentally given a weight of a dynamic point in the geometric module, as the object is found to be ``Parked", and its feature points do not need further scrutiny.

\section{Results and Discussion}
\label{Results}
In this section, the performance of OCD SLAM is evaluated on various sequences of the KITTI dataset~\cite{KITTI,KITTI-RAW}. In this quantitative evaluation\footnote{All results reported in this section are average values of 25 independent executions of each algorithm with each dataset; in order to reduce the effect of associated randomness.}, OCD SLAM is compared with the original ORB-SLAM2 and other state-of-the-art SLAM methods. An ablation study is also conducted to demonstrate the effectiveness of the cross disparity module. Finally, the computational performance of the proposed method is reported. These computational experiments were conducted on an AMD Ryzen 7 7700X CPU with a GeForce RTX 4080 GPU and 32 GB of RAM.

As evaluation parameters, we use ATE (Absolute Trajectory Error) and RPE (Relative Pose Error). The absolute trajectory error measures the Euclidean distance between the estimated camera trajectory and the aligned ground-truth trajectory at each timestamp. Alternatively, the relative pose error evaluates the local drift—that is, the relative motion between consecutive frames—with respect to the ground truth, in both rotation and translation.

\subsection{Evaluation on KITTI Odometry Dataset} 
The KITTI odometry~\cite{KITTI} consists of 22 stereo sequences, of which 11 sequences have ground truth trajectories as well. These sequences comprise stereo image pairs of outdoor scenes covering both urban road and residential areas.  In the eleven sequences, Sequences ``0, 2, 3, 5, 6, 8, 10" contain moderately dynamic objects, while Sequences ``1, 4, 7, 9" contain higher numbers of moving cars and pedestrians.  

\begin{table}[thbp]
  \caption{Comparison of Absolute Trajectory Error (ATE) and Relative Pose Error (RPE) between ORB-SLAM2 and the proposed OCD SLAM across KITTI Odometry sequences. }
  \renewcommand{\arraystretch}{1.5}
  \begin{center}
    \begin{tabular}{c| c c c | c c c}
      \hline
      \hline
      Sequence          & \multicolumn{3}{p{3cm}}{\centering \textbf{ORB-SLAM2}} & \multicolumn{3}{p{3cm}}{\centering \textbf {OCD SLAM}} \\ \cline{2-7}
       & \makecell{ATE\\(m)} & \makecell{RPE \\($^{\circ}/f$)} & \makecell{RPE\\(m/f)} & \makecell{ATE\\(m)} & \makecell{RPE \\($^{\circ}/f$)} & \makecell{RPE\\(m/f)} \\ \hline
    00	&1.30	&\textbf{0.19}	&0.05	&\textbf{1.13}	&0.20	&0.05\\
    01	&10.96	&\textbf{0.04}	&0.06	&\textbf{8.68}	&0.05	&\textbf{0.05}\\
    02	&5.62	&0.09	&0.03	&\textbf{4.94}	&0.09	&0.03\\
    03	&\textbf{0.70}	&\textbf{0.05}	&0.03	&\textbf{0.70}	&0.06	&0.03\\
    04	&\textbf{0.19}	&0.04	&0.02	&0.21	&0.05	&0.02\\
    05	&\textbf{0.77}	&\textbf{0.07}	&0.02	&0.81	&0.08	&0.02\\
    06	&0.87	&\textbf{0.04}	&0.02	&\textbf{0.64}	&0.05	&0.02\\
    07	&0.53	&\textbf{0.06}	&0.03	&\textbf{0.46}	&0.07	&0.03\\
    08	&3.35	&\textbf{0.06}	&0.09	&\textbf{3.27}	&0.07	&0.09\\
    09	&3.60	&\textbf{0.06}	&0.03	&\textbf{2.98}	&0.07	&0.03\\
    10	&1.14	&\textbf{0.07}	&0.03	&\textbf{1.13}	&0.08	&0.03\\

      \hline
      \hline
    \end{tabular}
    \label{tab:orb_slam}
  \end{center}
\end{table} 
   
\begin{figure}[htp]
\centering
\subfloat[]{\label{fig:01_seq}{\includegraphics[scale=0.22]{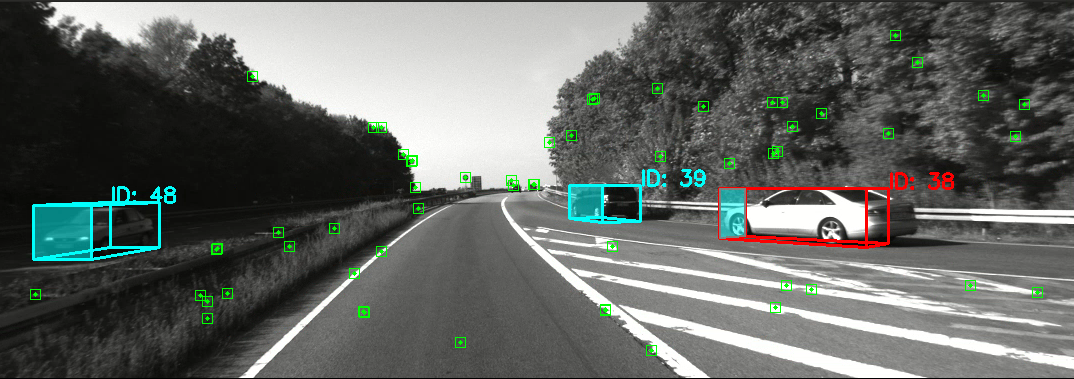}}}\hfill
\subfloat[]{\label{fig:07_seq}{\includegraphics[scale=0.22]{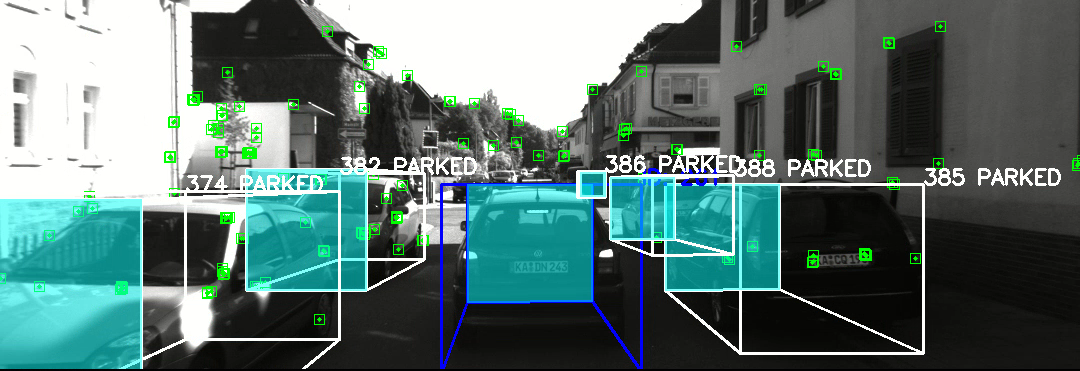}}}
\label{fig:odom_box}
\caption{The parked cars are enclosed by 3D boxes with white edges and labeled as PARKED, while the moving cars are shown with different colors.}
\end{figure}

Table~\ref{tab:orb_slam} presents the Absolute Trajectory Error (ATE) and Relative Pose Error (RPE) for ORB-SLAM2 and the proposed OCD SLAM on the KITTI odometry dataset. The columns $RPE \,(^{\circ}/f)$ and $RPE \,(m/f)$ denote the rotational and translational components of RPE, respectively. Results indicate that the proposed system achieves lower ATE values than ORB-SLAM2 in most sequences, except Sequences~03,~04, and~05. In particular, for highly dynamic sequences such as Sequences~01,~07, and~09, our method demonstrates significant improvements in absolute trajectory accuracy. Figures~\ref{fig:01_seq} and~\ref{fig:07_seq} illustrate the results of our method on Sequence~01 and~07, respectively. Both figures demonstrate that our method can distinguish between the static and dynamic vehicles; accordingly, it removes the features from the dynamic object while the static features remain intact. To avoid clutter, only the static points are shown in the figures.

It can also be observed that improvements in ATE do not consistently translate into lower RPE. This behavior is expected, since dynamic SLAM systems often discard or down-weight features associated with moving objects during bundle adjustment, thereby reducing long-term drift (improved ATE) but at the cost of showing higher short-term motion noise (similar or slightly worse RPE). Consequently, while the absolute trajectory is improved in our approach, the local odometry, which reflects relative position, the accuracy remains comparable to that of ORB-SLAM2, on an average. 

Table~\ref{tab:other_method} compares the proposed OCD SLAM with DynaSLAM~\cite{Dyna_SLAM}, the methods of Wen~et~al.~\cite{Shuhuan}, Yang~et~al.~\cite{YANG}, as well as RSO-SLAM~\cite{RSO_SLAM}. The evaluation is based on the Absolute Trajectory Error (ATE). ORB-SLAM2 and DynaSLAM results here have been reproduced using their \textit{open source code}, whereas for the remaining methods, the results have been taken directly from their original papers. The best result for each sequence is highlighted in bold.  
\begin{table}[th]
  \caption{Comparison of ATE of OCD SLAM with other state-of-the-art methods in KITTI Odometry sequences. }
  \begin{center}
    \begin{tabular}{c| c c c c c c}
      \hline
      \hline
      SEQ & \makecell{ORB \\SLAM2} &\makecell{Dyna \\ SLAM}	&\makecell{Wen\\ et al.} &\makecell{Yang\\ et al. }	&\makecell{RSO \\SLAM} &\makecell{OCD\\ SLAM}{\vspace{0.1cm}} 
\\ \hline 
    00	&1.30	&1.40	&1.31	&1.21	&1.27	&\textbf{1.13}\\
    01	&10.96	&9.40	&\textbf{8.78}	&10.0	&9.8	&\textbf{8.68}\\
    02	&5.62	&6.70	&5.83	&\textbf{4.96}	&5.3	&\textbf{4.94}\\
    03	&0.70	&\textbf{0.60}	&0.772	&$-$	&\textbf{0.56}	&0.70\\
    04	&0.19	&0.20	&0.20	&0.2	&\textbf{0.18}	&0.21\\
    05	&0.77	&0.80	&0.8	&0.8	&\textbf{0.75}	&0.81\\
    06	&0.87	&0.80	&0.78	&\textbf{0.5}	&0.74	&0.64\\
    07	&0.53	&0.50	&0.521	&0.5	&0.5	&\textbf{0.46}\\
    08	&3.35	&3.50	&3.425	&$-$	&3.49	&\textbf{3.27}\\
    09	&3.60	&1.60	&2.89	&3.15	&\textbf{1.7} &2.98\\
    10	&1.14	&1.20	&\textbf{0.97}	&1	    &1.16	&1.13\\

      \hline
    Total &0/11   &1/11   &2/11   &2/11      &4/11  &5/11  \\ 
      \hline
      \hline
    \end{tabular}
    \label{tab:other_method}
  \end{center}
\end{table} 
Results show that the proposed system performs better than other dynamic SLAM approaches in several sequences, which suggests that our cross disparity based geometric constraint, combined with motion categorization based on object velocity measurement, demonstrate improvements in trajectory estimation.  

Results of DynaSLAM suggest that simply removing potentially dynamic regions is not always beneficial, since such regions may also contain static objects that provide reliable features.  Approaches such as those of Wen~et~al. and Yang~et~al., which attempt to recover static features within segmented dynamic regions, improve robustness to some extent but remain limited. In these cases, some features belonging to truly static objects may be lost due to motion constraints on the segmented region, while several features from dynamic objects may occasionally be misclassified as static, thereby degrading performance.  

RSO-SLAM achieves modest improvement by incorporating stronger geometric constraints, as reflected in the results. However, its performance is still constrained by the same limitations. In contrast, the proposed system benefits from explicitly distinguishing between static and dynamic objects by tracking them over time and estimating their velocities; \textit{preventing} features on static objects from losing their effect in the final evaluation. This strategy enables more reliable feature selection, resulting in superior overall accuracy.

\subsection{Evaluation on KITTI Raw Dataset}

 OCD SLAM is further evaluated on several sequences from the KITTI raw dataset~\cite{KITTI-RAW}. These sequences present more challenging environments compared to the KITTI odometry dataset. For example, Sequence 0926-0013 contains both parked and moving cars, while Sequences 0926-0009, 0926-0014, 0926-0051, and 0926-0101 include highly dynamic traffic with a large number of moving vehicles. Sequences such as 0929-0004 and 1003-0047 represent heavy traffic with slow-moving vehicles.

These sequences also include objects that are not explicitly detected by the proposed 3D object detection module, such as buses, trucks, and trams. By using these data, we evaluated the effectiveness of the cross disparity module in handling such challenging cases.
\begin{table}[thb]
  \caption{Comparison of ATE of OCD SLAM with other state-of-the-art methods in KITTI Raw sequences.}
  \renewcommand{\arraystretch}{1.5}
  \begin{center}
    \begin{tabular}{c| c c c c }
      \hline
      \hline
      Sequence & \makecell{ORB \\SLAM2}	&\makecell{Dyna \\SLAM }	&\makecell{AirDos}&\makecell{OCD\\ SLAM}
\\ \hline
    0926-0009	&0.86	&0.95	&0.87	&\textbf{0.77}\\
    0926-0013	&0.96	&1.00	&\textbf{0.73}	&0.86\\
    0926-0014	&2.10	&2.00	&\textbf{1.85}	&\textbf{1.81}\\
    0926-0051	&0.80	&0.90	&0.76	&\textbf{0.62}\\
    0926-0101	&4.81	&4.26	&6.21	&\textbf{3.69}\\
    0929-0004	&1.81	&1.78	&1.54	&\textbf{1.27}\\
    1003-0047	&20.99	&\textbf{6.00}	&22.09	&12.09\\
      \hline
    Total &0/7   &1/7   &2/7   &5/7   \\ 
      \hline
      \hline
    \end{tabular}
    \label{tab:other_method_raw}
  \end{center}
\end{table}

In this evaluation, the proposed method was compared against ORB-SLAM2, DynaSLAM~\cite{Dyna_SLAM}, and AirDos~\cite{AirDOS}. Note that in the KITTI raw dataset evaluation, though several studies (such as ~\cite{DynaSLAM2, Peiliang, Shuhuan}) reported ATE results, many of them either applied scaled alignment or used different camera calibration parameters, making direct comparison inconsistent. Therefore, only those methods providing open-source stereo implementations were considered for evaluation. Table~\ref{tab:other_method_raw} reports the Absolute Trajectory Error (ATE) for these methods. Results indicate that our method consistently outperformed ORB-SLAM2 across all sequences. Compared with DynaSLAM and AirDos, the proposed method achieved better results in the majority of sequences. However, in Sequence 1003-0047, although our method improved upon ORB-SLAM2, it performed worse than DynaSLAM, which achieved the lowest error in this case. 
\subsection{Ablation Experiment}
In this section, each component of the proposed method is evaluated separately to analyze its contribution to the overall performance. Table~\ref{tab:ablation_study} presents results for three variants: using only the SMOKE module, only the cross disparity module, and also the full OCD SLAM system that integrates both modules.

\begin{figure*}[htp]
\centering
\begin{subfigure}{0.35\textwidth}
     \includegraphics[width=\textwidth]{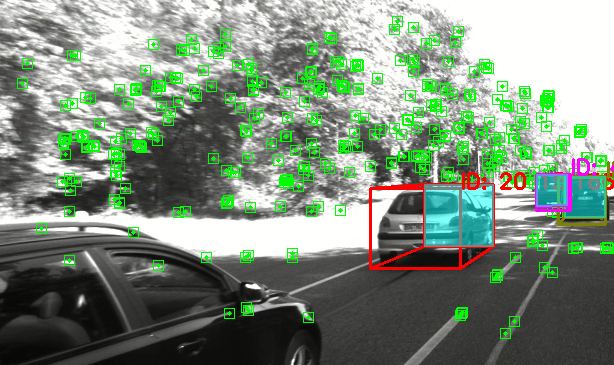}
     \caption{}
     \label{fig:0004_O_seq}
 \end{subfigure}
 \begin{subfigure}{0.35\textwidth}
     \includegraphics[width=\textwidth]{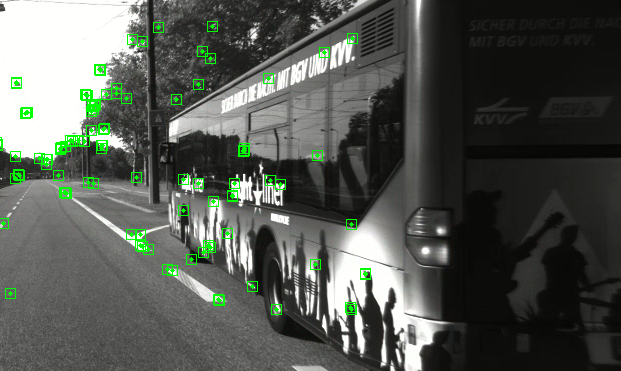}   
     \caption{}
     \label{fig:0101_O_seq}
 \end{subfigure}
 \hfill
 
 \begin{subfigure}{0.35\textwidth}
     \includegraphics[width=\textwidth]{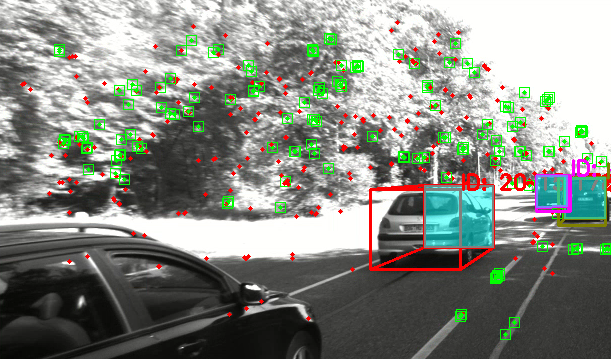}   
     \caption{}
     \label{fig:0004_C_seq}
 \end{subfigure}
 \begin{subfigure}{0.35\textwidth}
     \includegraphics[width=\textwidth]{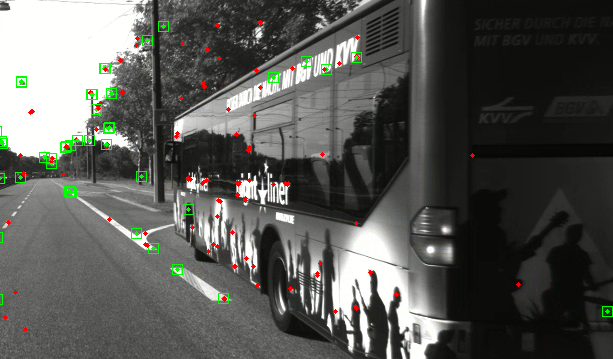}   
     \caption{}
     \label{fig:0101_C_seq}
 \end{subfigure}
  \caption{Ablation study illustrating the effect of different modules on dynamic feature removal. (a–b) Results using only the SMOKE module, where static features remain on moving objects (e.g., cars and buses) due to incomplete object detection. (c–d) Results using the cross disparity module, which detects dynamic features (shown in red).}
  \label{fig:abblation}
\end{figure*}

When using only the SMOKE (3D object detection) module, features on static objects (e.g., parked cars) are retained, while features on dynamic objects are removed. However, since 3D object detection can occasionally miss certain objects or fail to identify dynamic categories that are not predefined in the detection model, residual dynamic features may remain and negatively influence pose estimation. Figures~\ref{fig:0004_O_seq} and~\ref{fig:0101_O_seq} illustrate such cases, where features persist on cars and buses, respectively.
\begin{table}[thb]
  \caption{Ablation results of ATE on the KITTI Raw dataset comparing three configurations: SMOKE only, cross disparity only, and the full OCD SLAM system. }
  \renewcommand{\arraystretch}{1.5}
  \begin{center}
    \begin{tabular}{c| c c c}
      \hline
      \hline
      Sequence &\makecell{Only\\SMOKE} &\makecell{Only \\Cross Disparity}	&\makecell{OCD\\ SLAM (full)}
\\ \hline
    0926-0009	&0.85	&0.80	&\textbf{0.77}\\
    0926-0013	&0.99	&0.88	&\textbf{0.86}\\
    0926-0014	&1.91	&1.85	&\textbf{1.81}\\
    0926-0051	&0.84	&0.69	&\textbf{0.62}\\
    0926-0101	&4.46	&3.84	&\textbf{3.69}\\
    0929-0004	&1.72	&1.33	&\textbf{1.27}\\
    1003-0047	&12.81	&21.04	&\textbf{12.09}\\
	  \hline
      \hline
    \end{tabular}
    \label{tab:ablation_study}
  \end{center}
\end{table}
By contrast, when using only the cross disparity module (see figures~\ref{fig:0004_C_seq} and~\ref{fig:0101_C_seq}), dynamic features\footnote{A minor proportion of features on `categorically' known static objects, like trees, appearing as dynamic will be inconsequential; because the purpose here is to ensure the absence of static features on moving objects, and not the other way round.} can be identified effectively without relying on object categories, which leads to better performance in several sequences. For example, in Sequences~0926-0101 and~0929-0004, the cross disparity module achieved lower ATE values compared to SMOKE. On the other hand, cross disparity alone fails in scenarios with many slow-moving dynamic objects, as seen in Sequence~1003-0047 (in Table \ref{tab:ablation_study}), with significantly higher errors. 

\Mycommand{
A comparison between the results of DynaSLAM~\cite{Dyna_SLAM} (Table~\ref{tab:other_method_raw}) and those obtained using only the cross disparity module also indicates that the proposed cross disparity based geometric constraint provides better trajectory estimation performance compared to the depth-based approach in DynaSLAM.}

However, the full OCD SLAM system, which combines SMOKE and cross disparity modules, consistently yields the best results across all of the sequences. This demonstrates that the two modules are complementary: SMOKE ensures robust handling of categorically known dynamic objects, while cross disparity captures undetected or uncategorized moving objects, leading to improved trajectory estimation overall.  

\subsection{Computational Time}
Average computational time costs of the three main components of OCD SLAM — feature extraction, static point detection, and 3D object detection — are given in Table~\ref{time_table}. 
\begin{table}[htb]
    \centering
    \caption{Time costs of components of OCD SLAM.}
    \begin{tabular}{l c}
    \hline
    \hline
        \textbf{Task Component} & \textbf{Time per frame (ms)} \\
    \hline 
 Feature Extraction & 15.63 \\
 Static Point Detection & 38.97 \\
 3D Object Detection & 22.54 \\
 \hline
 \hline
    \end{tabular}
    \label{time_table}
\end{table}

In this result, the static point detection component was observed to consume the largest portion of computational time, followed by the 3D object detection module. Compared to ORB-SLAM2, which operates at approximately 23–25 frames per second (fps), the proposed method achieves a processing rate of 12–14 fps. Although this performance is somewhat lower for real-time operation, it can possibly be enhanced by reducing the overall inference time through code optimization and also better hardware components. 


\section{Conclusion}
\label{conclusion}
This paper proposes OCD-SLAM, a dynamic visual SLAM framework designed to improve camera localization and mapping in highly dynamic environments. The system extends ORB-SLAM2 by integrating SMOKE, a 3D object detection module, and a Kalman filter-based velocity estimation framework, enabling object-level motion classification. A novel geometric module is introduced here that defines a quantity named ``cross disparity'' and compares it with the disparity itself, establishing both temporal and stereo perspectives to identify dynamic features belonging to undetected moving objects. By combining velocity- and geometry-based motion filtering, the system successfully reduces the effect of dynamic features on the bundle adjustment process while preserving reliable static points.

Experimental evaluations on multiple sequences of the KITTI odometry and KITTI raw datasets demonstrate that the proposed method significantly improves absolute trajectory accuracy compared to (conventional) ORB-SLAM2 and other state-of-the-art dynamic SLAM approaches. The ablation study further showed that the inclusion of the cross disparity module provides a better detection approach in cases where the object detection module fails to detect uncategorized or partially occluded objects. In the detection of dynamic features and objects, the cross disparity module was observed to exhibit a complementary performance with other existing approaches (like SMOKE), making a compelling case for their deployment in combination.

In the current implementation, the 3D object detection module is trained to detect a very limited set of objects, such as cars, bicycles, and pedestrians. While this is sufficient for most scenarios in the KITTI Odometry dataset, the KITTI Raw dataset contains several additional object types that fall outside these categories. Although the proposed cross disparity module is useful in identifying dynamic features belonging to undetected objects, its effectiveness may suffer if too many such dynamic objects appear simultaneously in the view. Future work may strive to expand the set of detectable object categories to improve awareness and ensure more accurate motion estimation in complex real-world scenes.

\bibliographystyle{IEEEtran} 
\bibliography{IEEEabrv,ref}

\end{document}